\documentclass{article}
\usepackage{amsmath,graphicx,mlspconf}

\usepackage{subcaption} 
\usepackage{algorithm}
\usepackage{algorithmic}
\usepackage[bookmarks=false,draft]{hyperref}
\usepackage{mathtools}

\usepackage{booktabs}
\usepackage{soul} 
\usepackage{amsfonts,amssymb,amsbsy}
\usepackage{url}
\usepackage{xcolor}
\usepackage{footnote}
\usepackage{caption}
\usepackage{multirow}
\usepackage{tikz}
\usepackage{placeins}
\usepackage[titletoc]{appendix}

\hypersetup{
     colorlinks   = true,
     citecolor    = black,
     hidelinks     = true
}

\usepackage{microtype}

\newcommand{\tr}{^{\text{T}} }
\newcommand{\mb}[1]{\mathbf{#1}}

\newcommand{\partiald}[2]{\frac{\partial {#1}}{\partial {#2}}}

\newcommand{\cov}[1] { \mathrm{cov} {\left({#1}\right)}}

\newcommand{\x}{\mathbf{x}}
\newcommand{\f}{\mathbf{f}}
\newcommand{\y}{\mathbf{y}}
\newcommand{\X}{\mathbf{X}}

\newcommand{\id}{\text{d}}
\newcommand{\NormalNP} { \mathrm{N} }

\DeclareMathOperator*{\argmin}{arg\,min}

\usepackage{xspace}
\newcommand{\acr}[1]{\textsc{#1}\xspace}
\newcommand{\gp}{\acr{gp}}
\newcommand{\bo}{\acr{bo}}
\newcommand{\lcb}{\acr{lcb}}

\newcommand{\vbo}{\acr{vbo}}
\newcommand{\ep}{\acr{ep}}
\newcommand{\dbo}{\acr{dbo}}
\newcommand{\adbo}{\acr{adbo}}
\newcommand{\gps}{\acr{gp}{\!\,}s }

\newcommand{\mcmc}{\acr{mcmc}}
\newcommand{\mnd}{\acr{mnd}}

\title{CORRECTING BOUNDARY OVER-EXPLORATION DEFICIENCIES IN BAYESIAN OPTIMIZATION WITH VIRTUAL DERIVATIVE SIGN OBSERVATIONS}

\name{Eero Siivola$^1$, Aki Vehtari$^1$, Jarno Vanhatalo$^2$, Javier Gonz{\'a}lez$^{3}$, Michael Riis Andersen$^1$}
\address{$^1$Aalto University, Dept. of Computer Science, $^2$University of Helsinki,\\ Dept. of Math. and Stat., and Dept. of Biosciences, $^3$Amazon.com\thanks{Correspondence to \texttt{eero.siivola@aalto.fi} \newline Work done while Javier Gonz{\'a}lez was at the University of Sheffield}} 

\begin{document}

\maketitle

\begin{abstract} 
Bayesian optimization (\bo) is a global optimization strategy designed to find the minimum of an expensive black-box function, typically defined on a compact subset of $\mathcal{R}^d$, by using a Gaussian process (\gp) as a surrogate model for the objective. Although currently available acquisition functions address this goal with different degree of success, an over-exploration effect of the contour of the search space is typically observed. However, in problems like the configuration of machine learning algorithms, the function domain is conservatively large and with a high probability the global minimum does not sit on the boundary of the domain.
We propose a method to incorporate this knowledge into the search process by adding virtual derivative observations in the \gp at the boundary of the search space.
 We use the properties of \gps to impose conditions on the partial derivatives of the objective.   
The method is applicable with any acquisition function, it is easy to use and consistently reduces the number of evaluations required to optimize the objective irrespective of the acquisition used. We illustrate the benefits of our approach in an extensive experimental comparison.

\end{abstract}

\begin{keywords}
Bayesian optimization, Gaussian process, virtual derivative sign observation. 
\end{keywords}

\section{Introduction}\label{sec:intro}

Global optimization is a common problem in a very broad range of applications. Formally, it is defined as finding $\x_{\text{min}} \in \mathcal{X} \subset \mathcal{R}^d$ such that 
\begin{equation}\label{eq:problem}
\x_{\text{min}} = \argmin_{\x \in \mathcal{X}} f(\x),
\end{equation}
where $\mathcal{X}$ is generally considered to be a compact set of a Euclidean space. In this work, we focus on cases in which $f$ is a black-box function whose explicit form is unknown and that it is expensive to evaluate.
Thus, the goal is to locate $\x_{\text{min}}$ within a finite and typically small number of evaluations, which transform the original \emph{optimization} problem in a sequence of \emph{decision} problems.

Bayesian optimization of black-box functions using Gaussian Processes (\gps) as surrogate
priors has become popular in recent years (see, e.g. \cite{shahriari2016taking}). Treating the decision of where to evaluate the function $f$ next as a statistical \emph{inference} problem has been proven effective. This is typically using an acquisition function that balances \emph{exploration} and \emph{exploitation}.

A common problem that has not been systematically studied in the \bo literature is the tendency of most acquisition strategies to over-explore the boundary of the function domain $\mathcal{X}$. This issue is not relevant if the global minimum may lie on the border of the search space but in most cases, including when the search space is unbounded, this is not the case \cite{shahriari2016unbounded}. This effect has also been observed in the active learning literature and it is known to appear when the search is done myopically, as it is the case in most acquisitions functions \cite{ICML-2007-KrauseG}. Non-myopic approaches in \bo can potentially deal with this problem but they are typically very computationally expensive~\cite{GonzalezOL16}. 

\begin{figure}[bt!]
\centering
\includegraphics{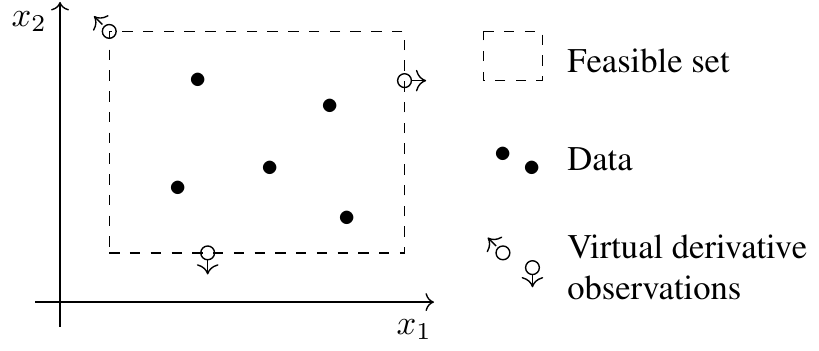}
\vspace{-0.8\baselineskip}
\caption{The concept of virtual derivative observations in two dimensional space. The arrows indicate the direction of the observed gradient. As there are no local minima on the borders, the gradient of the function always has a non-zero component pointing outwards from the feasible set. On the corners, both partial derivatives are non-zero.
\label{fig:augmentation} }
\end{figure}

In this paper we propose a new approach to correct the \emph{boundary over-exploration effect} of most acquisitions without increasing the computational overhead of currently available non-myopic methods. We demonstrate that when the local minimum is known not to lie on the boundary of $\mathcal{X}$, this information can be embedded into the model of the objective function. The assumption that $x_{\text{min}}$ does not sit on the boundary of the domain implies that the gradient of the underlying function points away from the  centre on the boundary. This property of the function can be incorporated into the model using \emph{virtual derivative observations}. Virtual derivative observations are unobserved data about the partial derivatives of the function that are treated similarly as true observations. In other words, we add pseudo derivative observations to the training set to induce the desired behaviour of the function at the boundary. As the derivative of a \gp is also a \gp, including virtual derivative observations in \gps is feasible with standard inference methods \cite{OHagan:1992}. In this work, we demonstrate that this reduces the number of required function evaluations giving rise to a battery of more efficient \bo methods. The concept of augmenting data with virtual derivative observations is illustrated in Figure \ref{fig:augmentation}.

\begin{figure}[bt!]
\centering
\setlength{\tabcolsep}{1pt}
\begin{tabular}[c]{cc}
\begin{subfigure}{0.18\textwidth}
    \centering
    \includegraphics{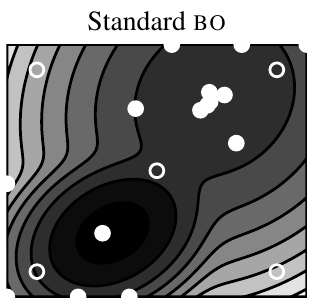}
	\vspace{-1.7\baselineskip}	    
    \caption{~}
	\label{fig:GPe}
\end{subfigure} &
\begin{subfigure}{0.18\textwidth}
	\centering
    \includegraphics{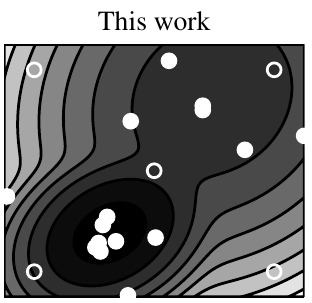}
	\vspace{-1.7\baselineskip}    
    \caption{~}
	\label{fig:GPed}
\end{subfigure}
\end{tabular}
\vspace{-1.0\baselineskip}
\caption{Sequence of 15 evaluations when optimizing a combination of Gaussians (darker colors represent lower function values) with (a) standard \bo (b) and the proposal of this work. The five white open circles are the points used to initialise the \gp. The 15 white balls are the acquisitions. The \gp-\lcb acquisition function was used in both cases (see Section \ref{seq:acquisitions} for details). With the new proposal, fewer evaluations are spent in the boundary, and more points are collected around the global optimum.
\label{fig:GPex}}
\end{figure}

The unwanted boundary over-exploration effect of the regular \bo is illustrated in Figure \ref{fig:GPex} (a). A simple function consisting of two Gaussian components is optimized with the standard \bo and the proposal of this work. The correction of the over-exploration effect is evident.

\subsection{Related Work}
Derivative observations have been used before in the \bo and \gp context to find minimum energy path
transitions of atomic rearrangements and to decrease the number of observations needed for finding the function optimum \cite{Koistinen_etal2016,wu2017bayesian}.

They can also be used to provide shape priors.
To constrain a function to have a mode in a specified location, virtual observations of first
derivative being zero and the second derivative being negative can be used \cite{gosling2007nonparametric}.
Virtual derivative observations where only the sign of the derivative is known can be used to add monotonicity
information \cite{riihimaki2010gaussian}.
To handle inference for the non-Gaussian contribution of the
derivative sign information, rejection sampling, expectation
propagation (\ep), and Markov chain
Monte Carlo (\mcmc) have been used \cite{gosling2007nonparametric,riihimaki2010gaussian,wang2016estimating}.

Surprisingly, over-exploration of boundaries has not been systematically studied before. A naive approach is to use a quadratic mean function to penalize the search in the boundary. However, this has strong limitations when the optimized function is multimodal or far away from being quadratic \cite{riihimaki2010gaussian}.

\subsection{Contributions}

The main contributions of this work are:
\begin{itemize}
\item A new approach for Bayesian optimization that corrects the over-exploration of the boundary of most acquisitions. The method is simple to use, can be combined with any acquisitions and always work equally or better than the standard approach. After a review of the needed background, the method is described in Section \ref{sec:method}.
\item A publicly available code framework\footnote{Framework available at \url{https://github.com/esiivola/vdsobo}} that contains an efficient implementation of the methods described in this work.
\item A comprehensive analysis of the performance of the proposed method in a variety of scenarios that should give the reader a precise idea about the (i) the loss in efficiency incurred in standard methods due to the boundary over-exploration and (ii) how this issue is significantly relieved with our proposal.
\end{itemize}

In addition to the previous points Section \ref{sec:conclusion} contains the main lessons learned in this work.

\section{Background and Problem Set-Up} \label{sec:theory}

The main iterative steps of any \bo algorithm are: (i) Model the objective function with \gp prior, which is updated based on the current set of function evaluations. (ii) Use an acquisition function, that depends on the posterior for the objective function, to decide what the next query point should be. Next, we visit both of them and detail how information from derivative observations can be naturally incorporated in the loop.

\subsection{Standard \gp Surrogate for Modeling of $f$}

At iteration $n+1$, we assume that we have evaluated the objective function $n$ times providing us the data $D =\{y^{(i)},\x^{(i)}\}_{i=1}^n$ where $y^{(i)}$ is, the possibly noisy, function evaluation at input location $\x^{(i)}$. To combine our previous knowledge about $f$ with the dataset $D$, we use a \gp to model $f$. In particular, a \gp prior is directly specified on the latent function with prior assumptions encoded in the covariance function $k(\x^{(1)}, \x^{(2)})$, which specifies the covariance of two latent function values $f(\x^{(1)})$ and $f(\x^{(2)})$. A zero mean Gaussian process prior 
\begin{equation} \label{eq:gp}
p(\f) = \NormalNP(\f | \mb{0}, \mb{K}),
\end{equation}
is chosen, where $\mb{K}$ is a covariance matrix between $n$ latent values $\f$ at input used for training, $\X = \left( \x^{(1)}, \ldots, \x^{(n)} \right)$, s.t. $\mb{K}_{ij} = k(\x^{(i)}, \x^{(j)})$.

In regression, $n$ noisy observations $\y$ and $o$ latent function values $\f_*$ at the test inputs $\X_*$ are assumed to have a joint Gaussian distirbution. With the noise variance $\sigma^2$, the covariance between the latent values at the training and test inputs $\mb{K}_*$, the covariance matrix of the latent values at the test inputs $\mb{K}_{**}$ and $n$ dimensional identity matrix $\mb{I}$, the joint distribution of the observations and latent values at the test inputs is
\begin{equation} \label{eq:gj}
\left[\begin{matrix}\y \\ \f_*\end{matrix} \right]  \sim \NormalNP \left( \mb{0}, \left[\begin{matrix} \mb{K} + \sigma^2 \mb{I} & \mb{K}_* \tr \\ \mb{K}_* & \mb{K}_{**}  \end{matrix}  \right]  \right).
\end{equation} Using the Gaussian conditioning rule, the predictive distribution for $\f_*$ can easily be computed

and the predictive distribution of the \gp can be written explicitly for any point in the domain.

\subsection{Acquisition Policy}\label{seq:acquisitions}
 
In this work, we concentrate on the \textit{lower confidence bound} (\lcb) acquisition function that minimizes the regret over the optimization area \cite{srinivas2009gaussian}. Although this is one of the most widely used acquisition function, it suffers from the over exploration effect described in the introduction of this work. As  we will detail later, the ability of \gps to handle derivative observations will be key to correct this effect.

\subsection{Incorporating Partial Derivative Observations in the Loop}\label{sec:derivatives}

Since the differentiation is a linear operator, the partial derivative of a (mean-square differentiable) Gaussian process remains a Gaussian process \cite{solak2003derivative}. Thus, using partial derivative values for prediction and making predictions about the partial derivatives at a given point is easy to incorporate in the model and in the \bo search. Since
\begin{align*}
\cov{ \partiald{f^{(i)}}{x^{(i)}_g}, \; f^{(j)} } &= \partiald{ }{ x^{(i)}_g}\; \cov{ f^{(i)}, \; f^{(j)}}, \\
\cov{ \partiald{ f^{(i)}}{ x^{(i)}_g}, \; \partiald{ f^{(j)}}{ x^{(j)}_h} } &= \frac{\partial^2 }{\partial x^{(i)}_g \partial x^{(j)}_h  }\; \cov{ f^{(i)}, \; f^{(j)}}
\end{align*}
covariance matrices in Equations \eqref{eq:gp} and \eqref{eq:gj} can be extended to include partial derivatives either as observations or as values to be predicted. 

Following Riihim{\"a}ki and Vehtari (2010) (\cite{riihimaki2010gaussian}), denote by $m \in \left\lbrace -1, 1\right\rbrace$ the partial derivative value in the dimension $j$ at $\tilde{\x}$. Then the probability of observing partial derivative is modelled using probit likelihood with a control parameter $\nu$ 
\begin{equation}
p\!\left( m \middle| \frac{ \partial \tilde{f}}{\partial \tilde{x}_{j}} \right) = \Phi \! \left( \frac{\partial \tilde{f}}{\partial \tilde{x}_{j}} \frac{m}{\nu} \right)\!,\text{ where  } \Phi(z)\! = \!\! \int_{- \infty} ^{z} \!\!\!\!\!\!\! N(t\,|\,0,1) \id t.
 \label{eq:probit}
\end{equation}

 Let $\bf{m}$ be a vector of $q$ partial derivative values at $\mb{\tilde{X}} = \left( \mb{\tilde{x}}^{(1)}, \ldots, \mb{\tilde{x}}^{(q)}  \right)$, $\mb{j}$ be a vector of the dimensions of the partial derivatives and $\bf{\tilde{f}}$ be the vector of latent values at $\bf{\tilde{X}}$ and let the partial derivatives of latent values be 
 $ \mb{\tilde{f}}' $.
Assuming conditional independence given the latent derivative values, the likelihood becomes
\begin{equation*}
p(\mb{m} \left. \middle| \right. \mb{\tilde{f}}') = {\prod_{i=1}^q}  \boldsymbol{\Phi} \left( \frac{\partial \mb{\tilde{f}}^{(i)}}{\partial \mb{\tilde{x}}^{(i)}_{ \mb{j}^{(i)}}} \frac{m^{(i)}}{\nu} \right).
\end{equation*}

With function values at $\X$ and partial derivative values at $\mb{\tilde{X}}$, the joint prior for $\bf{f}$ and $\bf{\tilde{f}}$ then becomes \begin{equation*}
\begin{aligned}
&p\left(\left[ \begin{matrix} \mb{f} \\ \mb{\tilde{f}}' \end{matrix} \right]\,  \middle| \, \left[\begin{matrix} \X \\ \mb{\tilde{X}}\end{matrix} \right]\right)   =    \NormalNP \left(\left[ \begin{matrix} \mb{f} \\ \mb{\tilde{f}}'\end{matrix} \right]\, \middle| \, \mb{0}, \left[ \begin{matrix} \mb{K_{f, f}} & \mb{K_{f, \tilde{f}'}}  \\ \mb{K_{ \tilde{f}',f}} & \mb{K_{ \tilde{f}', \tilde{f}'}} \end{matrix} \right]\right).
\end{aligned}
\end{equation*}
The joint posterior for the latent values and the latent value derivatives can be derived from the Bayes' rule
\begin{equation}
\begin{aligned}
&p(\mb{f}, \mb{\tilde{f}}'|\,\y, \mb{m}, \X, \mb{\tilde{X}})\! =\! \frac{p(\mb{f}, \mb{\tilde{f}}'|\, \X, \mb{\tilde{X}})p(\y \,|\,\mb{f})p(\mb{m}\,|\,\mb{\tilde{f}}')}{Z},
\end{aligned}
\label{eq:posterior}
\end{equation}
with $Z=\int\! p(\mb{f}, \mb{\tilde{f}}'| \X, \mb{\tilde{X}})p(\y \,|\, \mb{f})p(\mb{m} \,|\, \mb{\tilde{f}}') \id \mb{f} \id \mb{\tilde{f}}'.$ Note that since $p(\bf{m}\,|\,\mb{\tilde{f}}')$ is not Gaussian, the full posterior is analytically intractable and some approximation method must be used. We use expectation propagation (\ep) for fast and accurate approximative inference \cite{riihimaki2010gaussian}.

Model comparison is often done with the energy function, or negative log marginal posterior likelihood of the data
$E(\mb{y}, \mb{m}| \mb{X}, \tilde{\mb{X}}) = - \log p(\mb{y}, \mb{m}| \mb{X}, \tilde{\mb{X}})$.
If we are interested in only some part of the model, selected points $\left\{ \y^*, \mb{X}^* \right\}$ can be used to evaluate the model fit
\begin{equation}
\begin{aligned}
&E(\y^* |\mb{X}^*, \mb{y}, \mb{m}, \mb{X}, \tilde{\mb{X}})\! =\! -\! \log \frac{p(\y^*, \mb{y}, \mb{m} | \mb{X}^*, \mb{X}, \tilde{\mb{X}})}{p(\mb{y}, \mb{m} | \mb{X}, \tilde{\mb{X}})} .
\end{aligned}
\label{eq:energy}
\end{equation}

\section{Bayesian Optimization With Derivative Sign Observations}\label{sec:method}

In this section we illustrate how virtual derivative observations can be added to the edges of the search space. 
In essence, we use the same model as described in Section \ref{sec:derivatives} but where the derivative observations are replaced by virtual `observations' at the boundaries of the domain to correct for the described over-exploration effect.

\begin{figure}[bt!]
\centering
\setlength{\tabcolsep}{1pt}
\begin{tabular}[c]{cc}
\begin{subfigure}{0.23\textwidth}
	\centering
    \includegraphics{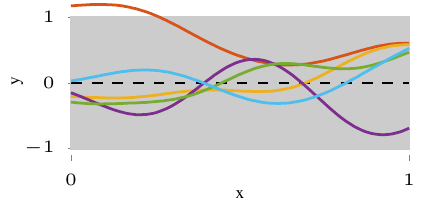}
	\vspace{-1.7\baselineskip}	
	\caption{~}
	\label{fig:GPprior}
\end{subfigure} &
\begin{subfigure}{0.23\textwidth}
	\centering
    \includegraphics{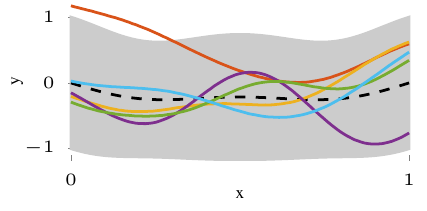}
    \vspace{-1.7\baselineskip}
	\caption{~}
	\label{fig:GPderprior}
\end{subfigure}

\end{tabular}
\vspace{-0.6\baselineskip}
\caption{The \gp prior in \bo visualized (a) without virtual derivative observations, (b) with virtual derivative observations on borders. Black dotted line is the mean of the \gp, the light gray area is 68\% central posterior interval and the five lines are random function samples from the prior. 
\label{fig:GPpriors}}
\end{figure}

\subsection{Virtual Derivative-Based Search}

To encode the prior information that the minimum is not in the boundary set, we propose the following dynamic approach. Just like in the regular \bo with \gp prior presented in the Section \ref{sec:theory}, the objective function is given a \gp prior which is updated according to the objective function evaluations so far. The next evaluation point is the acquisition function maximum, but if it is closer than threshold $\epsilon_b$ to the border of the search space, the point is projected to the border and a virtual derivative observation is placed at that point instead. After having added this virtual observation, the \gp posterior is updated and new proposal for the next acquisition is computed.
Algorithm \ref{alg:GPD} contains pseudocode for the proposed method.

\begin{algorithm}[tb]
\caption{Pseudocode of the proposed \bo method. The inputs are the acquisition $a$, the \emph{stopping criterion} and the \gp model. Note that this algorithm reduces to standard \bo when lines 4-6 are removed.
\label{alg:GPD}}
\begin{algorithmic}[1]
   \WHILE{\emph{stopping criterion} is False}
   \STATE Fit \gp to the available dataset $\X, \y$.
   \STATE Optimise acquisition function, $a$, to find select new location $\x$ to evaluate.
   \IF{$\x$ is close to the edge}
   \STATE Augment $\X$ with a virtual derivative sign observation at $\tilde{\x}$.
   \ELSE
   \STATE Augment $\X$ with $\x$ and evaluate $g$ at $\x$.
   \ENDIF
   \ENDWHILE
\end{algorithmic}
\end{algorithm}

 As there are no local minima on the borders, the gradient of the function always has a non-zero component pointing outwards from the feasible set. We have no knowledge if there are gradient components in other directions or about the magnitude of the gradient. Thus we only add a component pointing away from the feasible set, $\mathcal{X} \subset \mathcal{R}^d$ and use an observation model that only takes into account if the sign is positive or negative. If the feasible set is a hyper-cube, we can use partial derivatives as observations and thus avoid adding information about other directions.
As the control parameter $\nu$ in Equation \eqref{eq:probit} approaches 0 (and $m=1$),  the probit likelihood approaches the unit step function. This means that the likelihood values are close to $1$ for all partial derivative values $f' > 0$ and close to zero for all $ f' < 0$. Thus we can fix $m^{(i)}_{d_i} = \pm 1$ to positive and negative gradients. The effect of adding virtual derivative observations on the borders of a function is visualized in the Figure \ref{fig:GPpriors}. From the Figure it can be seen that the virtual derivative observations alter the \gp prior to resemble our prior belief of the location of the minimum.

Another parameter to be chosen is the threshold $\epsilon_b$. As acquisitions closer than the threshold value are always rejected, $\epsilon_b$ should not be too large. Another argument to avoid too large values is the fading information value of the virtual observations. If the \gp allows rapid changes in the latent values, virtual derivative observations affect the posterior distribution only very little.
Let $l$ be the diameter of the search space. Our experiments suggest that $\epsilon_b \approx 0.01 \cdot l$ is a good value in most applications.

\subsection{Adaptive Search}
For some practical applications, we might want to make the presented algorithm more robust to local minima on the border.
The following modifications to Algorithm \ref{alg:GPD} can be used.

Before placing a virtual derivative observation on the border, it can be checked whether or not the existing data supports the virtual gradient sign observation to be added to the model. This can be done by checking the energy values (Equation (\ref{eq:energy})) of virtual observations of different derivative values, $m_{d_i}^{(i)} \in \{-1, 1\}$.

Since virtual gradient observations only contain information about the sign of the partial derivative, they do not reduce the local variance of a \gp similarly as regular observations. As a result, if there are minima on the border, acquisitions might be proposed to locations where virtual observations already exists. If this happens, it is reasonable to remove the virtual observation before adding the new acquisition.

\section{Experiments\label{sec:experiments}}
In this section we introduce the four case studies performed to gain insight about the performance of the proposed method. First the details of the experiments are presented and then each study and its results are presented.  

\subsection{Experimental Set-Up}
The proposed Bayesian optimization algorithm was implemented in GPy toolbox\footnote{Toolbox available at: \url{https://sheffieldml.github.io/GPy/}}. We use zero mean \gp prior for regular observations and probit likelihood with $\nu=10^{-6}$ for the virtual derivative observations and the squared exponential covariance function for the \gp.%
Initial acquisitions are generated with a full factorial design with $2^d$ points (see 5.3.3.3. from \cite{croarkin2013engineering}).

Three \bo algorithms are used in the case studies. Standard \bo algorithm (referred as vanilla \bo, \vbo), algorithm with virtual derivative sign observations (referred as derivative \bo, \dbo) and adaptive version of \dbo (referred as \adbo). For the last two of these, virtual derivative sign observations are added if the next proposed point is within $1\%$ of the length of the edge of the search space to any border. For \adbo, old virtual derivative observations are removed before adding regular observation if the Euclidean distance between the points is less than $1\%$ of the length of the edge of the search space. 

\begin{figure}[tb!]
	\centering
    \includegraphics{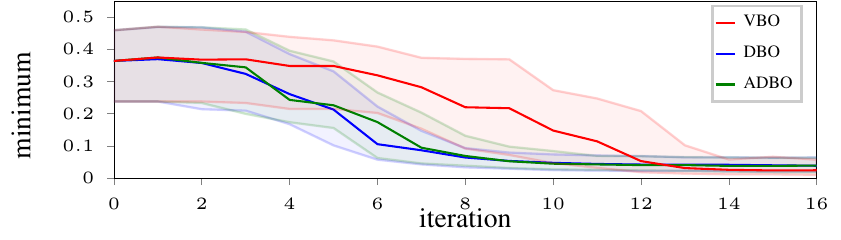}
    \vspace{-1.7\baselineskip}
\caption{Median and 25 and 75 percentiles of found minimum of 100 optimization runs as a function of iterations for \vbo, \dbo and \adbo. Optimization runs are performed for 3 dimensional \mnd-functions with additive noise of level $s=0.1$ and \lcb as an acquisition function. \label{fig:gaussianPerf}}
\end{figure}

\subsection{Case Study 1: A Simple Example Function}

The algorithm is used to illustrate the unwanted boundary over-exploration effect of the regular \bo. To show this, a simple function consisting of two Gaussian components is optimized with \vbo and \dbo using \lcb as an acquisition function. The function and 15 first acquisitions are visualized in Figure \ref{fig:GPex}, in the introduction.
The results show that \vbo over-explores the borders.

\subsection{Case Study 2: Random Multivariate Normal Distribution Functions \label{sec:gaussian}}
The algorithms are used to find the minimum of 100 different 3-dimensional multivariate normal distribution (\mnd) functions where the means and covariances are generated at random.

To mimic real life observations, Gaussian noise $\epsilon \sim N(0,0.1)$ is added to the observations $y(\x) = g(\x) + \epsilon$. 
25, 50, and 75 percentiles of found minimum values for the \mnd functions as a function of iterations are illustrated in Figure \ref{fig:gaussianPerf}.

The results show that performances of \dbo and \adbo are better than or equal to the performance of \vbo.
It can also be seen that the variance of the optimization performance between different optimization runs is smaller for \dbo and \adbo than for \vbo.

\begin{figure}[tb!]
	\centering
    \includegraphics{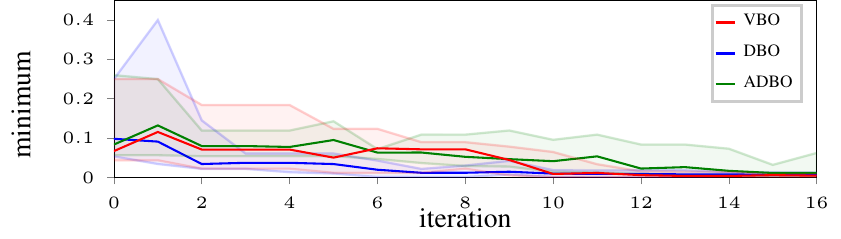}
    \vspace{-1.7\baselineskip}
\caption{Same as in Figure \ref{fig:gaussianPerf}, but with functions from Sigopt-library. \label{fig:sigoptPerf}}
\end{figure}
\begin{figure}[tb!]
\vspace{-0.8\baselineskip}
	\centering
    \includegraphics{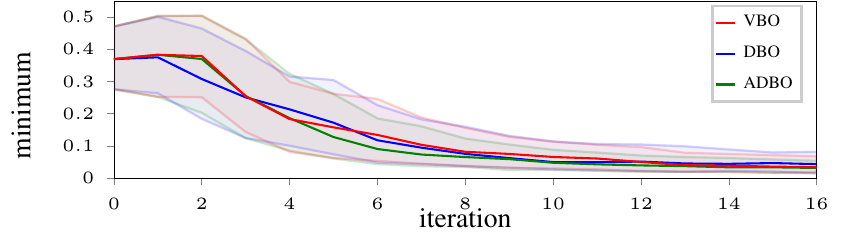}
    \vspace{-1.7\baselineskip}
\caption{Same as in Figure \ref{fig:gaussianPerf}, but with \mnd -functions that have local minimum on the edge of the search space. \label{fig:bordergaussianPerf}}
\end{figure}

\subsection{Case Study 3: Sigopt Function Library}
A benchmark function library\footnote{Function library available at: \url{https://github.com/sigopt/evalset}} Sigopt is developed for evaluating \bo algorithms \cite{dewancker2016stratified}.
When taking into account only three dimensional non-discrete functions without local border minima, the library outputs 14 functions. As in the previous case study, to mimic real use cases, the function observations are corrupted with additive Gaussian noise $y(\x) \sim g(\x) + \text{N}(0,0.1)$.
25, 50, and 75 percentiles of found minimum values of these functions as a function of iterations are illustrated in Figure \ref{fig:sigoptPerf}.
The results are similar as for \mnd functions. \dbo and \adbo still perform better than \vbo.
Similarly as before, the variance of the optimization performance between different optimization runs is notably smaller for \dbo than for \vbo. \adbo performs similarly as \vbo. Since there are less functions per dimension, the overall variability in the results is bigger and the percentile curves are not as smooth.

\subsection{Case Study 4: Simple Gaussian Functions With Minima on the Border}
The algorithms are used to find minimum of similar Gaussian functions as in Section \ref{sec:gaussian}, with the difference that the global minima of each function is exactly on the border of the search space.
The purpose of this case study is to show what happens to the performance of the proposed method if the a priori assumption is violated.
25, 50 and 75 percentiles of found minimum values of these functions as a function of iterations are illustrated in Figure \ref{fig:bordergaussianPerf}
As expected, the results show that \dbo does not perform as well as \vbo and \adbo.
Interestingly \dbo performs almost as well as \vbo, which shows the robustness of the proposed approach and makes it an appropriate `default' choice in most problems.

\subsection{Case Study 5: Hyper-parameter Optimization of RMSprop}

\begin{figure}[tb!]
	\centering
	\includegraphics{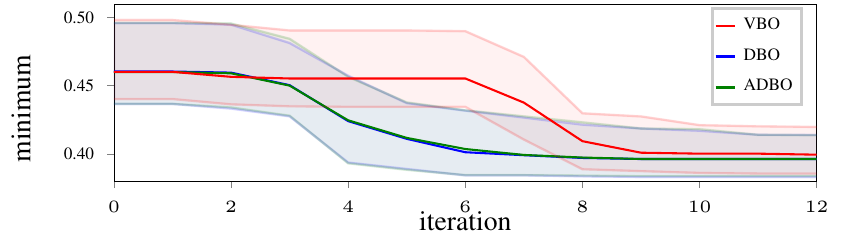}
\vspace{-1.7\baselineskip}
\caption{Same as in Figure \ref{fig:gaussianPerf}, but for optimizing hyper-parameters of a gradient descend algorithm and without adding noise. Validation error for the found minimum is displayed on y-axis. \label{fig:nn}}
\end{figure}

To show the performance for real data, the proposed algorithm was used to tune hyper-parameters of the RMSprop algorithm\footnote{RMSprop is an unpublished but established gradient descend method proposed by Geoff Hinton in \url{http://www.cs.toronto.edu/~tijmen/csc321/slides/lecture_slides_lec6.pdf}} that used in training a neural network for CIFAR10-data\footnote{Dataset available at: \url{https://www.cs.toronto.edu/~kriz/cifar.html}}. All the three presented optimizers are used to select the learning rate and decay of the RMSprop-algorithm. Classification error of the validation set as a function of iterations for 100 runs are illustrated in Figure \ref{fig:nn}. The results show that both the proposed methods perform better than \vbo.

\section{Conclusions} \label{sec:conclusion}
We have presented here a Bayesian optimization algorithm which utilizes qualitative prior information concerning the objective function on the borders. Namely, we assume that the gradient of the underlying function points towards the centre on all borders. Typical uses of Bayesian optimization concern expensive functions and in many applications qualitative knowledge of the generic properties of the function are known prior to optimization. 

The proposed \bo method has proved to significantly improve the optimization speed and the found minimum when comparing the average performance to the performance of the standard \bo algorithm without virtual derivative sign observations. The difference in performance is more significant if the assumption of non-existent global or local minima on the border of the search space holds, but is still notable if the assumption is relaxed so that the global minimum is not located on the border.

\section{Acknowledgments} 
We thank Juho Piironen and Kunal Ghosh for helpful comments to improve the manuscript. We also acknowledge the computational resources of the Aalto Science-IT project.

\bibliographystyle{IEEEbib}
\bibliography{bovobs}
\end{document}